\documentclass[letterpaper, 10 pt, conference]{ieeeconf}  

\IEEEoverridecommandlockouts                              

\usepackage{bm}
\usepackage{amsmath,amsfonts,amssymb,dsfont} 
\usepackage{multirow} 
\usepackage{graphicx} 

\usepackage{hyperref}
\hypersetup{
    colorlinks=true,
    linkcolor=blue,
    filecolor=magenta,      
    urlcolor=blue,
}

\usepackage{bm,amsmath,amsfonts,amssymb,dsfont} 
\usepackage{multirow} 
\usepackage{graphicx} 

\newcommand{\NewR}{\ensuremath{\mathds{R}}}
\newcommand{\mat}[1]{{\ensuremath{{\mathbf{#1}}}}}

\usepackage{setspace}

\title{\LARGE \bf
Single Shot 6D Object Pose Estimation
}

\author{Kilian Kleeberger$^{1}$ and Marco F. Huber$^{1,2}$
\thanks{$^{1}$Kilian Kleeberger is with the Department Robot and Assistive Systems and Marco F. Huber is with the Center for Cyber Cognitive Intelligence (CCI), Fraunhofer Institute for Manufacturing Engineering and Automation IPA,
        Nobelstra{\ss}e~12, 70569 Stuttgart, Germany
        {\tt\small \{kilian.kleeberger, marco.huber\}@ipa.fraunhofer.de}}%
\thanks{$^{2}$Marco F. Huber is with the Institute of Industrial Manufacturing and Management IFF, University of Stuttgart,
        Allmandring~35, 70569 Stuttgart, Germany
        {\tt\small marco.huber@ieee.org}}%
}

\begin{document}

\renewcommand*{\arraystretch}{1.4}

\maketitle

\begin{abstract} 
In this paper, we introduce a novel single shot approach for 6D object pose estimation of rigid objects based on depth images. For this purpose, a fully convolutional neural network is employed, where the 3D input data is spatially discretized and pose estimation is considered as a regression task that is solved locally on the resulting volume elements. With 65~fps on a GPU, our Object Pose Network (\mbox{OP-Net}) is extremely fast, is optimized end-to-end, and estimates the 6D pose of multiple objects in the image simultaneously. Our approach does not require manually 6D pose-annotated real-world datasets and transfers to the real world, although being entirely trained on synthetic data. The proposed method is evaluated on public benchmark datasets, where we can demonstrate that state-of-the-art methods are significantly outperformed.

\end{abstract}


\section{Introduction} 
Knowing the pose of objects is a crucial prerequisite for many robotic grasping and manipulation tasks. 6D object pose estimation (OPE) is a long-standing challenge and an open field of research since the early days of computer vision.
The pose of an object is fully described by a translation vector $\mat{t} \in \NewR^3$ and a rotation matrix ${\mat R \in \mathrm{SO}(3)}$ of its body-fixed coordinate system relative to a given reference frame. The task of 6D OPE is challenging due to the variety of objects in the real world,
sensor noise, clutter and occlusion in the scene and varying lighting conditions, which affect the appearance of the objects in the image. Object symmetries, which are common in man-made and industrial environments, result in pose ambiguities and have to be addressed.

\begin{figure}[tb]
\centering
\includegraphics[scale=0.26]{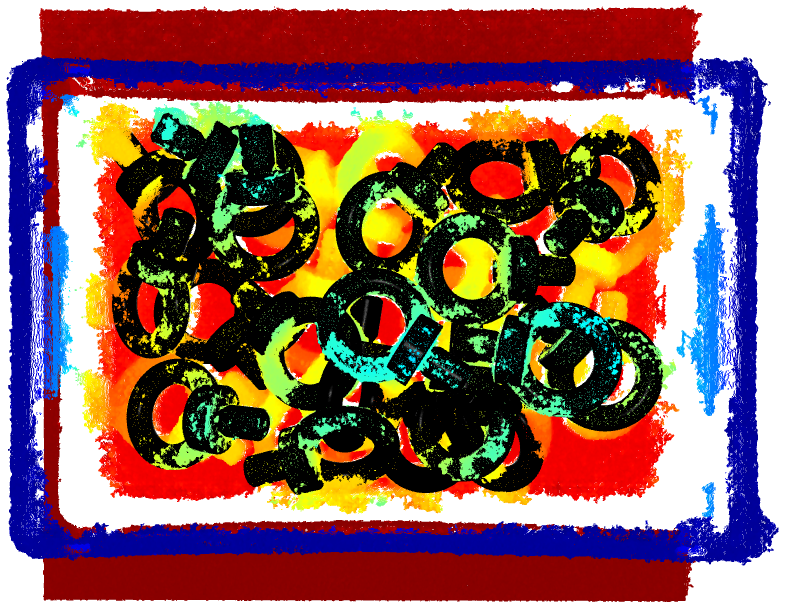}
\caption{Object pose estimates (black) of our approach on real-world data for ring screws after ICP refinement.
Although our models are fully trained on synthetic data, they successfully transfer to the real world.}
\label{fig:result_real_world_data}
\end{figure}


In recent years, research in 6D OPE has been dominated by convolutional neural network (CNN) based approaches. The state-of-the-art approaches solve this as a classification problem, in which the pose space is discretized into bins and a CNN
is used to predict a pose bin~\cite{SSD-6D, Sock, OPE_Classification, V&K, RenderforCNN, ECCV2018bestpaperaward, PoseRBPF}. The pose space, however, is continuous and thus, we solve the 6D OPE problem by means of a regression task.
Recent works regress the
2D image coordinates of the object's 3D bounding box
and use a P$n$P algorithm to estimate the object's 6D pose~\cite{BB8, Tekinetal., PVNet}. Because of the more intuitive interpretation, this task seems
easier to learn than directly predicting the pose. To the best of our knowledge, we are the first directly predicting the 6D object pose in terms of position and angles for the orientation as a regression task in a single shot framework, without requiring any computational overhead like P$n$P, clustering, or post-processing steps like the iterative closest point (ICP) algorithm for pose refinement. We propose a solution for dealing with discrete and revolution object symmetries, which need to be considered to avoid inconsistent loss signals during training. Contrary to most other methods using RGB or RGB-D images as input, we use depth data only. Our approach is
inspired by~\cite{YOLOv1, GraspDetection_Redmon} and uses a fully convolutional architecture to process the depth image.

Approaches for 6D OPE relying on a prior semantic segmentation step~\cite{BB8, PoseCNN, 6D_OPE_IROS2018, IROS2018_PoseInterpreterNetworks, PVNet} cannot be used for scenes of many parts of the same type in bulk. When the parts overlap, they require instance segmentation, which is hard to obtain for highly occluded and cluttered scenes. Additionally, the quality of the segmentation masks also influences the quality of the pose estimates.
As shown in~\cite{YOLOv2, SSD}
for 2D object detection, a single shot system reasons globally on the image and the whole pipeline is optimized end-to-end, thus being both faster and more accurate compared to multi-stage systems~\cite{FasterR-CNN,MaskRCNN}.


So far, direct pose regression methods have
been less
successful than classification~\cite{SSD-6D, 3DPoseRegression, ECCV2018bestpaperaward}. A major reason for this is the lack of precisely annotated training data 
without data redundancy.
Usual datasets show a limited variability of poses relative to the camera due to numerous image acquisitions of the same scene from different views and are therefore less suitable for precise regression tasks~\cite{datasetTLess, datasetTejanietal., datasetDoumanoglouetal., datasetDesk3D, Hinterstoisser_ADD/ADI_LINEMOD+, datasetBrachmann}.
Because creating and annotating datasets with 6D poses is time-consuming and does not scale, since the process has to be repeated for every new application, we tackle this problem via simulation. This enables generating large-scale datasets with flawless annotations. 
We use sim-to-real transfer techniques to allow the model generalizing in the real world.

We benchmark our proposed method on the Sil\'{e}ane~\cite{RomainBrégier} and Fraunhofer~IPA~\cite{IROS_2019_dataset_paper} datasets, which are challenging due to multiple instances of the same object type and a high amount of clutter and occlusion. By using our data generator to generate training data, we significantly outperform the state-of-the-art methods
without requiring additional post-processing steps like ICP or duplicate removal.
%

In summary, the main contributions of this work are:
\begin{itemize}
\item A novel single shot approach for 6D OPE that is highly robust to occlusions between objects and does not rely on any post-processing to get accurate results, even on low resolution input images ($128 \times 128$ pixel)
\item Two novel loss functions that can properly deal with object symmetries and formulate the 6D OPE task as a regression instead of classification problem
\item A framework for training the pose estimator entirely in simulation and enabling it to generalize in the real world without requiring 6D pose-annotated real-world training data

\end{itemize}


The paper is structured as follows. In the next section, related work is reviewed. In Section~\ref{sec:approach} the proposed approach is described. Experimental evaluations are provided in Section~\ref{sec:evaluation}. Pros and cons of our approach are discussed in Section~\ref{sec:discussion}. The paper closes with a conclusion.

\section{Related Work} 

We first review related
work on classical feature and template matching methods before taking a closer look at newer CNN-based methods for 6D OPE using RGB, depth, or RGB-D images as input.

\subsection{Classical Approaches}
Traditionally, the problem of 6D OPE is tackled by template or feature matching.
In template-based methods~\cite{Hinterstoisser_ADD/ADI_LINEMOD+, Hinterstoisser_templatematching},
rigid templates are constructed by rendering the 3D object model from different viewpoints, which are then matched into the sensor data at different locations. A similarity score is used for the evaluation of the pose hypothesis. These methods are useful for detecting texture-less objects, but occlusions significantly reduce the
performance. This is due to the low similarity between the template and the sensor data if the object is occluded.
%
%
In feature-based methods, local features from the image are matched to the 3D object model to recover the 6D pose based on the spatial relationship~\cite{FeatureMatching, FeatureMatching_2, V&K}.
These methods are designed to handle changes in size, viewpoint and illumination. While being robust to occlusion and scene clutter, they can only reliably handle objects with sufficient texture.

\subsection{CNN-based Approaches}
In recent years, research in 6D OPE has been dominated by CNN-based approaches.
In~\cite{Sock}, a binning of the angles and the height for the 6D OPE in depth images is introduced. The accuracy and speed are limited due to the discretization and the multi-stage nature of the system.
\cite{3DPoseRegression} solves pose estimation as a regression instead of classification task, but focuses on the retrieval of the orientation only based on the object's bounding box.

Deep-6DPose~\cite{Deep-6DPose} extends Mask~R-CNN~\cite{MaskRCNN} with an additional branch for estimating the full 6D pose based on the given region proposal.
%
After a coarse segmentation, BB8~\cite{BB8} predicts the 2D projections of the corners of the object's 3D bounding box. The 6D pose is estimated
by using a P$n$P algorithm and a final CNN per object is trained to refine the pose. Due to the usage of multiple separated CNNs, the approach is not optimized end-to-end and is time-consuming for the inference.
PoseCNN~\cite{PoseCNN} decouples the translation and rotation estimation based on a prior semantic segmentation step and thus, also requires multiple stages.

SSD-6D~\cite{SSD-6D} extends the SSD~\cite{SSD} detection framework to 6D OPE by predicting the 2D bounding boxes with SSD, hypothesizing 6D poses from the network output (6D pose pool) and running a refinement step (ICP). The rotation estimation is treated as a classification problem by decomposing the 3D rotation space into discrete viewpoints and in-plane rotations.

PPR-Net~\cite{PPR_Net}, the winner method of the ``\textit{Object Pose Estimation Challenge for Bin-Picking}'' at IROS 2019\footnote{\url{http://www.bin-picking.ai/en/competition.html}}, uses PointNet++~\cite{PointNet++} and estimates a 6D pose for each point in the point cloud of the object instance to which it belongs.
Afterwards, density-based clustering in 6D space is applied and the final pose hypotheses are obtained by averaging the predicted poses for each identified cluster.

A recent result from the Benchmark for 6D Object Pose Estimation (BOP)~\cite{BOP} is that methods based on point-pair features currently perform best on various datasets, outperforming methods based on template matching, 3D local features, and machine learning. This is in contrast to many other computer vision tasks such as image classification~\cite{AlexNet, DenseNet}, object detection~\cite{YOLOv1, YOLOv2, SSD}, semantic segmentation~\cite{FullyConvolutionalNetworksforSemanticSegmentation}, instance segmentation~\cite{MaskRCNN}, etc. being dominated by deep learning. With this work, we advance the state-of-the-art for learning-based methods by providing a simple pipeline for an end-to-end trainable general purpose 6D pose estimator, which can be trained entirely on synthetic data and generalizes in the real world. 

\section{Object Pose Network}
\label{sec:approach}
\begin{figure*}[thpb]
\begin{footnotesize}
\centering
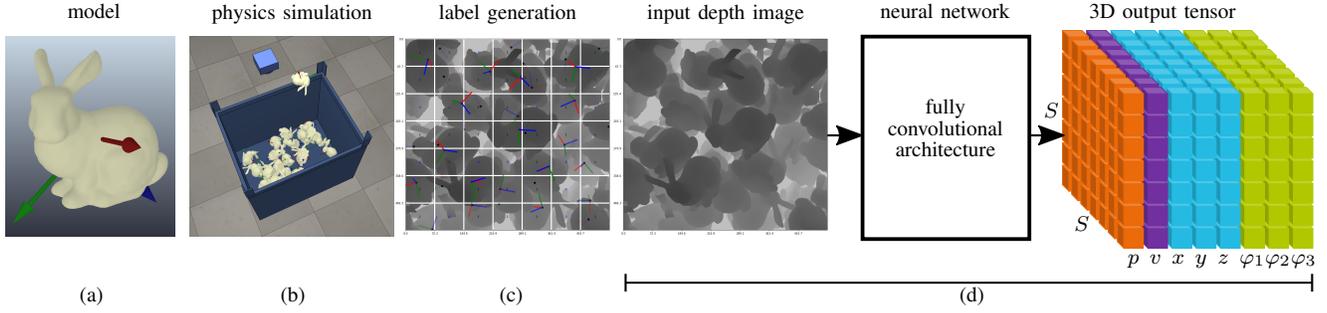
\caption{Overview:
(a) 3D object model with body-fixed coordinate system (can be a CAD or previously scanned model).
(b) Physics simulation for the generation of training data.
(c) Rendered depth image from the simulation with overlayed $S \times S$ grid. Based on the 3D position of the object coordinate system, the objects in the scene are assigned to the volume elements.
(d, left) Rendered depth image used as input for the neural network.
(d, middle) The proposed CNN architecture
based on a DenseNet-BC~\cite{DenseNet}.
(d, right) 3D output tensor of the neural network, where each spatial location consists of a vector comprising the probability $p$ that the cell contains an origin, visibility $v$, positions $x$, $y$, $z$, and angles $\varphi_1$, $\varphi_2$, $\varphi_3$. }
\label{fig:Overview}
\end{footnotesize}
\end{figure*}

In this section, we describe the process of data generation for training our neural network, the parameterization of the network's output, the loss function, the network architecture, the training procedure, and the technique for the
transfer of the model from simulation to the real world. Fig.~\ref{fig:Overview} shows an overview of our approach.

\subsection{Data Generation}
In order to generate the training data for \mbox{OP-Net},
we use a physics simulation that produces scenes of relevant scenarios as depicted in Fig.~\ref{fig:Overview}~(b). The depth images and segmentation masks are generated in
perspective
projection. The ground truth annotation consists of a class label, the 6D pose, i.e., translation vector $\mat{t}$ and rotation matrix $\mat R$ relative to the coordinate system of the 3D sensor, a segmentation label, and a visibility score
of each object instance
in the scene. The visibility $v \in [0,1]$ is the ratio between the number of visible pixels and the number of pixels without any occlusion.



\subsection{Parameterization of the Output}
The output of our model is inspired by the parameterization
proposed in~\cite{GraspDetection_Redmon, YOLOv1}. Instead of predicting (oriented) rectangles
in 2D, we estimate the 6D pose of the objects in the image. By discretizing the 3D scene in $S \times S$ volume elements, the 6D pose regression task is solved locally, i.e., individually for each volume element.
Each volume element
comprises an 8-dimensional
vector containing the probability $p$, visibility $v$, positions $x$, $y$, $z$, and angles $\varphi_1$, $\varphi_2$, $\varphi_3$ in a given convention.
For the ground truth generation, the pose of an object is assigned to a volume element
if the origin of the body-fixed coordinate system of the 3D object model is located within the volume element as visualized in Fig.~\ref{fig:Overview}~(c). If multiple origins
fall into the same volume element,
the object with higher visibility $v$
is assigned as ground truth. All volume elements not containing an object are filled with a zero vector.
The output of the network is a $S \times S \times 8$ tensor as depicted in Fig.~\ref{fig:Overview}~(d, right).

The probability channel
reflects how confident
the model is that the volume element contains an origin of a 3D object model. Both the probability
and visibility prediction at each spatial location are used to filter the results at test time.

The parameterization can naturally be extended to class probabilities of the objects. For instance, a variant of YOLO~\cite{YOLOv2} can differentiate between 9,000 object classes alongside with confidence scores and bounding box coordinates. 
Also an estimate of how well an object can be grasped, can naturally be added. Thus, the parameterization is general and can be used for any rigid 3D object.


\begin{figure*}[thpb]
\begin{footnotesize}
\centering
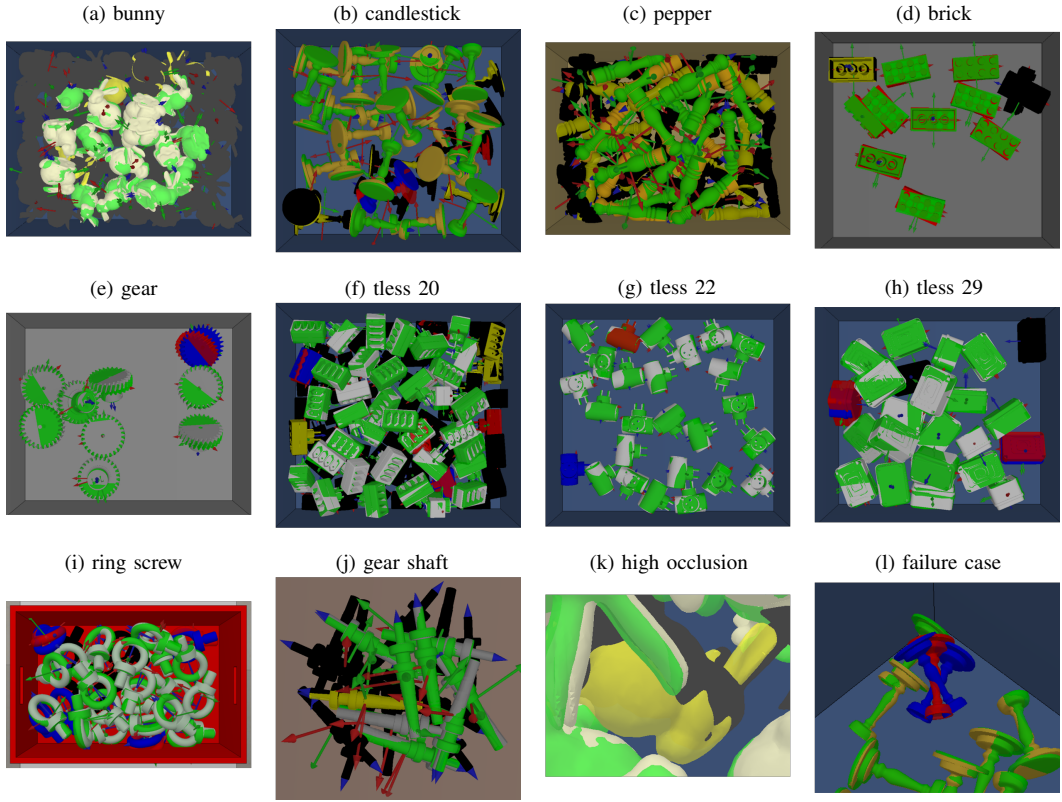
\caption{Some qualitative results on scenes from the
Sil\'{e}ane~\cite{RomainBrégier} (a--h) and
Fraunhofer~IPA~\cite{IROS_2019_dataset_paper} (i, j) datasets without any post-processing, a successful retrieval of an object with $v=0.2$ (k), and a failure case due to a limitation of the parameterization (l). All images show a 3D visualization of the ground truth and the result of our method in a simulation. Object instances that do not need to be found ($v\leq0.5$) are visualized in black, while especially some objects at the border of the 3D visualization are black because they are truncated in the depth image.
Objects that need to be found and whose pose was correctly estimated are visualized in their original color. Otherwise they are visualized in blue. Correctly found objects are visualized in green (true positives). Red indicates that the object is either a duplicate or too far off from the ground truth pose (false positive). Objects that were correctly found, but their pose was not of interest for retrieval (too much occluded) are displayed in yellow and are neither considered as true nor false positive.}
\label{fig:Results}
\end{footnotesize}
\end{figure*}


\subsection{Loss Functions}
\label{sec:lossfunctions}
To train the network, the multi-task
loss function
\begin{equation}
\mathcal{L}\!=\!\sum_{i=1}^{S^2}\!\bigg(\!\lambda_1 (p_i - \hat{p}_i)^2+\Big[\! \lambda_2 (v_i - \hat{v}_i)^2 + \lambda_3 \big( \mathcal{L}_\mathrm{pos} + \lambda_4 \mathcal{L}_\mathrm{ori} \big)\!\Big] p_i \!\bigg)
\end{equation}
%
%
is optimized, where a hat indicates estimates of the network. $\lambda_1$, $\lambda_2$, $\lambda_3$, and $\lambda_4$ are manually tuned weights for
the different loss terms. While $\lambda_1=0.1$, $\lambda_2=0.25$, and $\lambda_4$ are constant, $\lambda_3=8v^3$ is a function of the ground truth visibility $v$. The less an object is visible, the lower the loss of the pose error is weighted. The idea is to let the network focus on the well visible objects and alleviate convergence issues due to pose ambiguities because of occlusions.

To stabilize the training, the position $\mat{x}=[x, y, z]^\top$ of the object is estimated relative to the volume element, i.e. $x, y, z \in (0,1)$, while $z$ is the position between the near and far clipping plane of the 3D sensor. We use
\begin{equation}
\mathcal{L}_\mathrm{pos} = \lVert\mat{x}-\hat{\mat{x}} \rVert^2~,
\end{equation}
with $\lVert.\rVert$ being the $L^2$ norm.

We consider
two loss functions for the estimation of the orientation. First,
\begin{equation}
\mathcal{L}_\mathrm{ori1} = \lVert\bm{\varphi}-\hat{\bm{\varphi}}\rVert^2~,~\lambda_4=1~,
\end{equation}
with $\bm{\varphi}=[\varphi_1, \varphi_2, \varphi_3]^\top$.
Second,
\begin{equation}
\mathcal{L}_\mathrm{ori2} = \displaystyle\min\limits_{\mat{p}_{\mat R} \in \mathcal{R}_{\mat R}(\mathcal{P})} \lVert\mat{p}_{\mat R} - \hat{\mat{p}}_{\mat R}\rVert~,~\lambda_4=0.5~,
\end{equation}
with $\mat{p}_{\mat R}$ being the orientation part of the pose representative $\mat p \in \mathcal{R}$ of pose $\mathcal{P}$ and comprising the relevant axis vectors of the rotation matrix $\mat R$ depending on the proper symmetry group of the object, based on~\cite{RomainBrégier_Distance, RomainBrégier}.
This loss function computes the Euclidean distance between the prediction and the closest ground truth.
%
%
%
%
%
%
%
Since it gives better results, we bound the angles for both loss functions, i.e., $\varphi_1,\varphi_2 \in [0,2\pi)$ and $\varphi_3 \in [0,2\pi/k)$ are mapped to $[0,1)$
where $k \in \mathbb{N}$ represents the order of the cyclic symmetry.
For objects with a revolution symmetry, we omit the
output feature-map for~$\varphi_3$.
%
%
%
%
%
%
In Section~\ref{sec:EvaluationonSimulatedData} we compare the performance of the
loss functions for the regression of the angles.








\subsection{Network Architecture}
The input of our model
is a single normalized depth image in
perspective projection. The data is processed with a fully convolutional architecture and mapped to a 3D output tensor as shown in Fig.~\ref{fig:Overview} (d).
In the experiments, we use an input resolution of $128 \times 128$ pixel,
a DenseNet-BC~\cite{DenseNet} with 40 layers and a growth rate of 50, which represents the number of feature-maps being added per layer, and $S=16$. We choose a DenseNet-BC because it promotes gradient propagation by introducing direct connections between any two layers with the same feature-map size and has a high parameter efficiency.
The
network architecture consists of four dense blocks and downsampling is performed three times via $2 \times 2$ average pooling to reduce the size of the feature-maps from $128 \times 128$ to $16 \times 16$ and preserve the spatial information.
ReLU activation functions are employed in the dense blocks and sigmoid functions for
the 3D output tensor.
With this architecture, forward passes are performed with an average frame rate of 65~fps on a Nvidia Tesla V100.






\subsection{Training}
During training, the error
for the entire probability channel
and the error of the visibility and 6D pose
for the specific entries in the 3D output tensor that contain ground truth poses are backpropagated. Thanks to the utilized simulation, there is an abundant source of data that allows training the network from scratch. We use the Adam optimizer with an initial learning rate of 0.01, monitor the validation loss, reduce the learning rate by a factor of 10 if the loss did not improve for three epochs, and train the network
for about 50 epochs on the data generated by the physics simulation.


\subsection{Sim-to-Real Transfer}
To transfer the
model from simulation to the real world,
we use a technique called domain randomization~\cite{DomainRandomization}. We randomize various aspects of the simulation, e.g., pose and size of distractor objects that must be ignored by the network, and apply different augmentations with varying intesitiy to the rendered training images, e.g., adding noise, blurring, elastic transformations, dropout, etc. This allows the 6D pose estimator generalizing in the real world, although being trained entirely on synthetic data. The main advantage of this technique is that it
requires no samples
from the real world. This is in contrast to domain adaptation~\cite{GANs_Sim-to-Real, GANs_Pixel-LevelDA} techniques, which are usually based on Generative Adversarial Networks (GANs)~\cite{GANs}. Apart from being hard to train and often yielding fragile training results, GANs require the acquisition of (unlabeled) real-world samples for new applications,
which negatively impacts scalability.




\section{Experimental Evaluation}
\label{sec:evaluation}


We evaluate our single shot approach for 6D OPE on bin-picking scenarios. These consist of multiple rigid objects of the same type, which are stored chaotically in a bin.
6D OPE is challenging due to the cluttered scene with multiple and heavy occlusions.
We benchmark the performance of our method on the noisy data from the Sil\'{e}ane dataset~\cite{RomainBrégier} and the real-world data from the Fraunhofer~IPA dataset~\cite{IROS_2019_dataset_paper}. These show highly chaotic scenes typical for bin-picking. 
Fig.~\ref{fig:result_real_world_data} and Fig.~\ref{fig:Results} depict
qualitative results of \mbox{OP-Net}. 

\subsection{Evaluation Metric} 


A common evaluation metric for 6D OPE is ADD by Hinterstoisser et al.~\cite{Hinterstoisser_ADD/ADI_LINEMOD+}, which accepts a pose hypothesis if the average distance of model points between the ground truth and estimated pose is less than 0.1 times the diameter of the smallest bounding sphere of the object. As this metric cannot handle symmetric objects, ADI~\cite{Hinterstoisser_ADD/ADI_LINEMOD+} was introduced for handling those. The ADI metric is widely used, but can fail to reject false positives as demonstrated in~\cite{RomainBrégier}.
For evaluation,
we use the metric
provided by Br\'{e}gier et al.~\cite{RomainBrégier_Distance, RomainBrégier}, which is suitable for rigid objects, for scenes of many parts in bulk, and properly considers cyclic and revolution object symmetries.
A pose representative $\mat{p} \in \mathcal{R}(\mathcal{P})$ comprises a translation vector $\mat{t}$ and the relevant axis vectors of the rotation matrix $\mat R$ depending on the object's proper symmetry group.
The distance between a pair of poses $\mathcal{P}_1$ and $\mathcal{P}_2$ is defined as the minimum of the Euclidean distance between their respective pose representatives:
\begin{equation}
d(\mathcal{P}_1, \mathcal{P}_2) = \displaystyle\min\limits_{\mat{p}_1 \in \mathcal{R}(\mathcal{P}_1),~\mat{p}_2 \in \mathcal{R}(\mathcal{P}_2)} \lVert\mat{p}_1 - \mat{p}_2\rVert
\end{equation}
A pose hypothesis is accepted (considered as true positive) if the minimum distance to the ground truth is less than 0.1 times the object's diameter. Following~\cite{RomainBrégier}, only the pose of objects that are less than 50\% occluded are relevant for the retrieval.
The metric breaks down the performance of a method to a single scalar value named average precision~(AP)
by taking the area under the precision-recall curve.




\subsection{Comparison}
\label{sec:EvaluationonSimulatedData}
Depending on the object, the Sil\'{e}ane dataset~\cite{RomainBrégier} comprises 46 to 325 images and is too small for training deep neural networks. Therefore, we precisely rebuild the setup from the dataset in simulation.

To allow \mbox{OP-Net} generalizing on real-world data without being trained on real-world samples, we randomize the bin pose in the simulation during the data generation process
and augment the synthetic images during training.
As the real-world 3D sensor operates in perspective mode, the obtained point clouds are projected to perspective depth images
because it shows fewer pixels with missing depth information compared to the orthogonal projection. The missing values are interpolated before the image is fed into the neural network.


Table~\ref{table:Results_simulated_data} shows the performance of our method against the state-of-the-art in terms of the metric by Br\'{e}gier et al.~\cite{RomainBrégier_Distance, RomainBrégier}. We provide performance results for different loss functions for the regression of the angles.
%
%
%
\mbox{OP-Net} outperforms other classical~\cite{PPF, Hinterstoisser_ADD/ADI_LINEMOD+} and learning-based~\cite{Sock,PPR_Net}
approaches
without requiring post-processing steps~(PP) like duplicate removal
or ICP, even on low resolution input images.
%
%
%
Applying randomizations in the simulation and augmentations to the training data allows the model generalizing on real-world data without ever being trained on samples from the target domain. Additional
PP steps like duplicate removal and ICP further improve the results.

\begin{table*}[hbtp]$ $
\begin{tiny}
\caption{
Average precision (AP) values of different methods and comparison of the performance of different loss functions for our method (best results in bold).
%
%
The models are trained on the data generated with our physics simulation and all images from the Sil\'{e}ane~\cite{RomainBrégier} and Fraunhofer~IPA~\cite{IROS_2019_dataset_paper} datasets are used for testing.
Results marked with * are taken from the ``\textit{Object Pose Estimation Challenge for Bin-Picking}'' at IROS 2019.
}
\begin{spacing}{0.95}
\begin{center}
\begin{tabular}{l|c|c|c|c|c|c|c|c|c|c|}
object & bunny~\cite{RomainBrégier} & candlestick~\cite{RomainBrégier} & pepper~\cite{RomainBrégier} & brick~\cite{RomainBrégier} & gear~\cite{RomainBrégier} & tless 20~\cite{RomainBrégier} & tless 22~\cite{RomainBrégier} & tless 29~\cite{RomainBrégier} & ring screw~\cite{IROS_2019_dataset_paper} & gear shaft~\cite{IROS_2019_dataset_paper} \\
object symmetry based on~\cite{RomainBrégier_Distance,RomainBrégier} & (no proper & (revolution) & (revolution) & (cyclic, & (revolution) & (cyclic, & (no proper & (cyclic, & (cyclic, & (revolution) \\
& symmetry)  &              &              & $k=2$)  &              & $k=2$)  & symmetry)  & $k=2$) & $k=2$) & \\
\hline
\mbox{OP-Net} (ours) with $\mathcal{L}_\mathrm{ori1}$ & 0.92 & 0.94 & \textbf{0.98} & 0.41 & 0.82 & 0.85 & 0.77 & 0.51 & 0.88 & 0.99 \\
\mbox{OP-Net} (ours) with $\mathcal{L}_\mathrm{ori1}$ and PP & \textbf{0.94} & \textbf{0.97} & \textbf{0.98} & 0.42 & \textbf{0.84} & \textbf{0.88} & \textbf{0.86} & \textbf{0.58} & 0.93 & 0.99 \\
\mbox{OP-Net} (ours) with $\mathcal{L}_\mathrm{ori2}$ & 0.74 & 0.95 & 0.92 & 0.79 & 0.58 & 0.56 & 0.53 & 0.36 & 0.73 & \textbf{1.0} \\
\mbox{OP-Net} (ours) with $\mathcal{L}_\mathrm{ori2}$ and PP & 0.76 & 0.96 & 0.93 & \textbf{0.80} & 0.60 & 0.58 & 0.55 & 0.39 & 0.75 & \textbf{1.0} \\
\hline
PPF~\cite{PPF,RomainBrégier} & 0.29 & 0.16 & 0.06 & 0.08 & 0.62 & 0.20 & 0.08 & 0.19 & - & - \\
PPF PP~\cite{PPF,RomainBrégier} & 0.37 & 0.22 & 0.12 & 0.13 & 0.63 & 0.23 & 0.12 & 0.23 & - & - \\
LINEMOD+~\cite{Hinterstoisser_ADD/ADI_LINEMOD+,RomainBrégier} & 0.39 & 0.38 & 0.04 & 0.31 & 0.44 & 0.25 & 0.19 & 0.20 & - & - \\
LINEMOD+ PP~\cite{Hinterstoisser_ADD/ADI_LINEMOD+,RomainBrégier} & 0.45 & 0.49 & 0.03 & 0.39 & 0.50 & 0.31 & 0.21 & 0.26 & - & - \\
Sock et al.~\cite{Sock} & 0.74 & 0.64 & 0.43 & - & - & - & - & - & - & - \\
PPR-Net~\cite{PPR_Net} & 0.82 & 0.91 & 0.80 & - & - & 0.81 & - & - & \textbf{0.95}* & 0.99* \\
PPR-Net with ICP~\cite{PPR_Net} & 0.89 & 0.95 & 0.84 & - & - & 0.85 & - & - & - & - \\
\end{tabular}
\end{center}
\end{spacing}
\label{table:Results_simulated_data}
\end{tiny}
\end{table*}

\section{Discussion}
\label{sec:discussion}

In the following, we summarize strengths and discuss limitations of our approach.

\subsection{Strengths}
Due to
learning plausible object pose configurations,
\mbox{OP-Net} is highly robust to occlusions.
Fig.~\ref{fig:Results} (k) exemplarily shows a successful estimate
of an object that is barely visible in the input depth image.

For classical approaches and CNN-based methods
relying on region proposals~\cite{Deep-6DPose, SSD-6D, ECCV2018bestpaperaward, PoseRBPF, 3DPoseRegression}
or a prior segmentation step~\cite{BB8, PoseCNN, 6D_OPE_IROS2018, IROS2018_PoseInterpreterNetworks} for pose estimation,
the runtime increases linearly with the number of objects present in the scene, which is not optimal for very cluttered scenes. Our approach only requires a single feed-forward to estimate the 6D pose of
multiple
objects in the image simultaneously at 65~fps on a GPU.
PPR-Net~\cite{PPR_Net} requires 200~ms
(GTX 1060) for the forward pass and additional clustering in 6D space plus pose averaging afterwards. \mbox{OP-Net} is faster because of using a more compact parameterization and not requiring any post-processing steps.
Contrary to multi-stage methods,
single shot approaches allow a global reasoning
and still make accurate predictions locally.
We frame 6D OPE as a regression problem and are therefore not limited in the accuracy given by the
discretization of the pose space.

Our proposed loss functions are designed such that they can properly handle object symmetries. This avoids inconsistent loss signals during training causing convergence issues, because different orientations may generate identical observations. PoseCNN~\cite{PoseCNN} uses the ADD and ADI metric~\cite{Hinterstoisser_ADD/ADI_LINEMOD+} for symmetric objects as loss function, while ADI does not properly handle object symmetries
as shown in~\cite{RomainBrégier}.
Using the metric provided by Br\'{e}gier et al.~\cite{RomainBrégier_Distance} as loss function, which properly handles object symmetries and compares the retrieved object pose against every possible ground truth, allows an efficient pose distance computation contrary to ADD and ADI, which depend on the sampling of the model points and require an iteration over all model points.

Manually creating and annotating a dataset with 6D poses
is a very time-consuming
process that does not scale because it has to be repeated for every new application. Therefore, we tackle the problem via
simulation which allows to easily generate large-scale 6D pose-annotated datasets. We apply various augmentations to the synthetic images and show that our approach transfers to the real world, although being trained entirely on synthetic data.
As we train our network on a bunch of synthetic variations that generalize in the real world, we are also independent of the 3D sensor technology being used.
The Sil\'{e}ane~\cite{RomainBrégier} and Fraunhofer~IPA~\cite{IROS_2019_dataset_paper} datasets were recorded with different sensors.
In our experiments, we used the same network architecture, parameter configuration, and augmentation techniques which emphasises the scalability of our approach.

\subsection{Limitations}
The parameterization of the network output faces several limitations. If multiple objects fall into the same volume element, the object with higher visibility
is assigned as ground truth.
A failure case is exemplary visualized in Fig.~\ref{fig:Results}~(l), where two candlesticks, which have their origin in the center of mass, fall into the same volume element with their origin. Since this scenario can hinder better convergence behaviour during training, a possible solution would be to predict multiple 6D poses per volume element or to introduce a discretization in $z$-direction resulting in a 3D probability map
and therefore a 4D output tensor. A further limitation
is that the origin of the 3D object model has to be inside the image
in order to be detected by the proposed pose estimator.
In case the estimation of the 6D pose of those objects is of relevance, one could
add pixels at the border of the image
to include the origin in the image.
When the origin of the 3D object model is very close to the border of a volume element, the object might also get detected by neighboring volume elements. These can be filtered out via setting a higher threshold for $\hat{p}$ or an additional duplicate removal step based on the distance between the origins or pose representatives of the 6D object poses in the result.

\section{Conclusion and Future Work}
\label{sec:conclusion}

In this paper, we proposed a novel single shot approach for 6D object pose estimation of rigid objects based on depth images that is fast by design, end-to-end trainable, and facilitates direct 6D pose regression of the visible objects in the image simultaneously.
Our experiments demonstrate that our method
is highly robust to occlusions,
can handle symmetric objects, and provides accurate pose estimates, even in highly cluttered scenes.
\mbox{OP-Net} outperforms state-of-the-art methods on the Sil\'{e}ane
dataset~\cite{RomainBrégier} by a large margin without any post-processing steps,
even on low resolution input images.
The proposed 6D pose estimator can be transferred to the real world, although being entirely trained on synthetic images and annotations by randomizing the simulation and augmenting the synthetic images. In doing so, pose estimation becomes independent of the
3D sensor technology being used in the real world.
%
%
In future work, we will focus on giving a quality estimate for each predefined grasp pose on the 3D object model based on
grasping trials in a physics simulation.


\section*{Acknowledgment}
This work was partially supported by the Baden-W\"urttemberg Stiftung gGmbH (Deep Grasping -- Grant No. NEU016/1) and the Ministry of Economic Affairs of the state Baden-W\"urttemberg (Center for Cyber Cognitive Intelligence (CCI) -- Grant No. 017-192996). We would like to thank our colleagues for helpful discussions and comments.


\bibliographystyle{IEEEtran}
\bibliography{IEEEabrv,references}

\end{document}